\pdfoutput=1
\documentclass{llncs}
\usepackage{llncsdoc}
\usepackage{amsmath}
\usepackage{graphicx}
\usepackage{amsfonts}
\usepackage{amsmath}
\usepackage{subfig}
\usepackage{soul}
\usepackage{hyperref}
\usepackage{xcolor}
\usepackage{float}
\usepackage [english]{babel}
\usepackage [autostyle, english = american]{csquotes}
\usepackage{authblk}
\newcommand{\eg}{e.\,g.}
\newcommand{\ie}{i.\,e.}
\newcommand{\etal}{\emph{et al.}}

\begin{document}

\title{Self-supervised Dense 3D Reconstruction from Monocular Endoscopic Video}
\author{Xingtong Liu \inst{1} \and Ayushi Sinha \inst{1} \and Masaru Ishii \inst{2} \and Gregory D. Hager \inst{1} \and Russell H. Taylor \inst{1} \thanks{Russell H. Taylor is a paid consultant to and owns equity in Galen Robotics, Inc. These arrangements have been reviewed and approved by JHU in accordance with its conflict of interest policy.} \and Mathias Unberath \inst{1}}

\institute{The Johns Hopkins University, Baltimore, USA,\\ \email{xliu89@jh.edu}
\and
Johns Hopkins Medical Institutions, Baltimore, USA}

\maketitle

\begin{abstract}
We present a self-supervised learning-based pipeline for dense 3D reconstruction from full-length monocular endoscopic videos without \textit{a priori} modeling of anatomy or shading. Our method only relies on unlabeled monocular endoscopic videos and conventional multi-view stereo algorithms, and requires neither manual interaction nor patient CT in both training and application phases. In a cross-patient study using CT scans as groundtruth, we show that our method is able to produce photo-realistic dense 3D reconstructions with submillimeter mean residual errors from endoscopic videos from unseen patients and scopes.
\end{abstract}

\section{Introduction}
Minimally invasive procedures in the head and neck typically employ surgical navigation systems to provide surgeons with additional anatomical and positional information to avoid critical structures. Computer vision-based navigation systems that rely on the intra-operative endoscopic video stream and do not introduce additional hardware are both easy to integrate into clinical workflow and cost-effective, but require registration of pre-operative data, such as CT scans, to the intra-operative videos~\cite{sinha2018endoscopic}. For 3D-to-3D registration algorithms, estimating an accurate and dense intra-operative 3D reconstruction is necessary to ensure acceptable performance of the system. However, obtaining such reconstructions is not trivial due to problems such as textureless surface, specular reflectance, lack of photometric constancy across frames, and tissue deformation.

Several methods have been explored for 3D reconstruction in endoscopy. Multi-view stereo methods, such as Structure from Motion (SfM) and Simultaneous Localization and Mapping (SLAM)~\cite{leonard2018evaluation,grasa2014visual}, are able to reconstruct 3D structure and estimate camera poses in feature-rich scenes. However, the paucity of features in endoscopic images can cause these methods to produce sparse and unevenly distributed reconstructions, which may lead to inaccurate registration. Mahmoud \etal~propose a quasi-dense SLAM-based method~\cite{mahmoud2017slam}, which is still potentially sensitive to hyper-parameters. Shape from Shading based methods~\cite{goncalves2015perspective} model the relationship between appearance and depth but usually oversimplify the problem, which can result in inaccurate reconstructions in cases with, \eg, specular reflection. Hardware-based solutions~\cite{yang2016compact}, \eg~structured light camera, are still challenging because of non-Lambertian properties of tissues and the paucity of features. Deep-learning based methods have recently been explored to solve the single-frame dense 3D reconstruction task in monocular endoscopy. Simulation-based works use synthetic dense depth maps generated from patient-specific CT~\cite{mahmood2018deep} to solve the problem of unpaired data. A self-supervised method has been proposed by Liu~\etal~\cite{Liu2019Self} that only requires unlabeled endoscopic videos. To our knowledge, all existing deep-learning based methods in monocular endoscopy are based on single frames. How to effectively fuse predictions from endoscopic video frames to generate a full-length reconstruction has not been studied yet. However, it constitutes an important step for single-frame based methods to be useful in surgical navigation systems.

\begin{figure}[t]
	\centering
	\includegraphics[width=100mm]{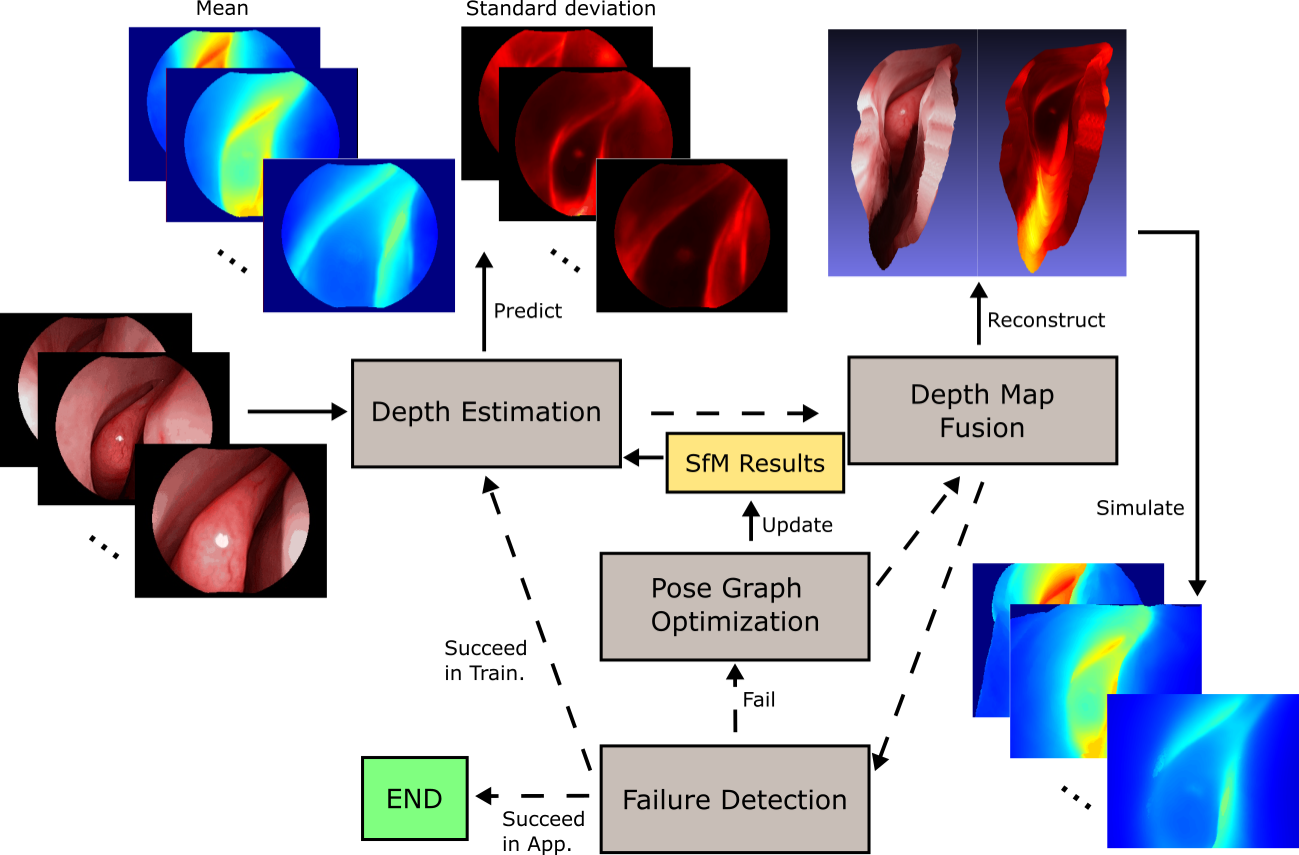}
	\caption{Overall pipeline. There are $4$ modules in both training and application phases, which are \emph{Depth Estimation}, \emph{Depth Map Fusion}, \emph{Failure Detection}, and \emph{Pose Graph Optimization}. \emph{Depth Estimation} is used to train a single-frame depth estimation network and generates depth maps with uncertainty estimates from an endoscopic video. \emph{Depth Map Fusion} produces dense 3D reconstructions with uncertainty estimates and simulates depth maps for training in \emph{Depth Estimation}. \emph{Failure Detection} determines whether a dense 3D reconstruction fails. If it does, \emph{Pose Graph Optimization} will optimize the estimated camera poses of all video frames from SfM. SfM will be rerun if a failure for the same video sequence happens again. In the training phase, the next iteration of the pipeline begins once all dense 3D reconstructions are generated and no failure cases remain. In the application phase, the pipeline starts with running SfM on the video to estimate camera poses and sparse point clouds and ends when the 3D reconstruction is generated successfully. The dotted arrows represent the executing order of the pipeline.}
	\label{fig:overall_pipeline}
\end{figure}


In this paper, we present a self-supervised approach for dense 3D reconstruction from full-length monocular endoscopic videos. Our contributions are as follows: (1) To the best of our knowledge, this is the first self-supervised method for 3D dense reconstruction from full-length monocular endoscopic videos. (2) We propose a bootstrapping pipeline with novel modules, namely \emph{Depth Estimation} that can estimate uncertainties in depth predictions,  \emph{Depth Map Fusion} that produces dynamic 3D reconstructions with uncertainty estimates, \emph{Failure Detection} that detects reconstruction failures automatically, and \emph{Pose Graph Optimization} that refines estimated camera poses from SfM. (3) We demonstrate that we are able to generate accurate and photo-realistic 3D reconstructions purely from unlabeled endoscopic videos from unseen patients and endoscopes.


\section{Methods}
\textbf{Overall Pipeline:} The overall pipeline, shown in Fig.~\ref{fig:overall_pipeline}, trains a single-frame depth estimation network iteratively in the training phase, and generates dense 3D reconstructions which are correct up to a global scale from full-length endoscopic videos in the application phase. 
\newline
\noindent\textbf{Depth Estimation with Uncertainty:} The depth estimation module is a single-frame depth estimation network based on the work by Liu~\etal~\cite{Liu2019Self}. Due to the ill-posed nature of single-frame depth estimation, uncertainty estimation is important to reconstruct accurate 3D models from videos, but it has not been explored in any previous learning-based depth estimation methods in monocular endoscopy that we are aware of. Inspired by Kendall~\etal~\cite{kendall2018multi}, besides a mean depth map obtained as in~\cite{Liu2019Self}, an additional standard deviation depth map is predicted by our network. We assume that the depth prediction of the corresponding pixel in the input video frame follows an independent Gaussian distribution that can be represented with a mean and standard deviation, which are $\mu_{i,j}$ and $\sigma_{i,j}$ for 2D location $(i, j)$. We minimize log-likelihood loss to maximize the joint probability of the training data from SfM, which are sparse point clouds and camera poses, by optimizing the distribution parameters; \ie~mean and standard deviation depth maps. Symbols that appear in~\cite{Liu2019Self} are briefly introduced in this work. We use four types of losses for network training. The \emph{Sparse Depth Loss} for frame $j$ is defined as 
\begin{equation}
    L_{\text{sd}}\left(j\right) = \dfrac{1}{\sum M_j}\sum { \left( M_j \left(\text{ln}\left(S_j + \epsilon\right) +  \dfrac{\left( Z_j^{\text{s}} - Z_j\right)^2}{2S_{j}^{2} + \epsilon} \right) \right)} \quad \text{, where}
\end{equation}
$M_j$, $Z_j^\text{s}$, $Z_j$, and $S_j$ are the sparse binary mask, sparse depth map, mean depth map, and the standard deviation depth map of frame $j$, respectively. 
We change the sparse mask $Z_j^\text{s}$ from soft to binary weights to force the network to predict high standard deviation values for regions in which it is uncertain. $\epsilon$ is used to avoid numerical instability. This loss ensures that the predicted depth maps agree with the sparse point clouds from SfM and gathers uncertainty information of sparse point clouds from SfM. 
We change the \emph{Depth Consistency Loss} of~\cite{Liu2019Self} from a normalized L$2$ loss to a dense log-likelihood loss. The loss is defined by replacing $M_j$ and $Z_j^{\text{s}}$ in \emph{Sparse Depth Loss} with $W_{k,j}$ and $\check{Z}_{k,j}$, respectively. $\check{Z}_{k,j}$ is the dense depth map warped from frame $k$ to frame $j$ using the \emph{Depth Warping Layer} of~\cite{Liu2019Self}, and $W_{k,j}$ is the intersection of valid regions of $Z_j$ and $\check{Z}_{k,j}$. This loss ensures that predicted depth maps from adjacent frames are geometrically consistent and gathers the uncertainty information of camera poses from SfM. 
The \emph{Sparse Flow Loss} of~\cite{Liu2019Self} is reused because of its robustness to outliers from SfM by jointly using the information from pair-wise feature matches and camera poses from SfM. 
The \emph{Dense Simulation Loss} is used after one iteration of the pipeline when dense 3D reconstructions are produced in \emph{Depth Map Fusion}. This loss is defined by replacing  $Z_j^{\text{s}}$ and $M_j$ in the \emph{Sparse Depth Loss} with $Z_j^{\text{sim}}$ and $\Tilde{M}_j$, which are simulated depth maps from 3D reconstructions and intersection of valid regions between $Z_j$ and $Z_j^{\text{sim}}$. By using simulated depth maps for network training, the network essentially learns the depth information of all frames in the video sequence in a condensed form. This loss, with \emph{Depth Consistency Loss}, overcomes the primary problem of SfM that there are few sparse points in textureless and deep regions. By combining all losses, the network learns what patterns in video frames tend to have higher depth uncertainties. In the application phase, all predicted depth maps are re-scaled using the \emph{Depth Scaling Layer} of~\cite{Liu2019Self} to make their scales consistent with the results from SfM for further processing.
\newline
\noindent\textbf{Adaptive Depth Map Fusion:} We extend a volumetric method based on the work by Curless~\etal~\cite{curless1996volumetric} for generating dynamic 3D reconstructions by fusing predicted depth maps from a single-frame depth estimation network. We replace the original signed distance function with the truncated signed distance function in~\cite{zach2007globally}. We propose to replace the original evidence accumulating method that ignores the temporal information with an adaptive exponential averaging method, which exploits both pixel-wise depth uncertainty information and temporal order, to fuse the depth information from \emph{Depth Estimation}. This change is based on two observations. First, the essential difference between depth measurements from depth sensors and a neural network lies in the error distribution of depth values. A sensor-based depth measurement of a static scene from a fixed camera pose approximately follows a Gaussian distribution, which means that we are able to get a more accurate depth estimation by simply averaging multiple depth measurements. However, this is not the case for a deep learning-based single-frame depth estimation method because it is an inherently ill-posed problem. A certain photometric pattern could correspond to multiple depth values, which is true even without considering global scale variation. Consequently, the network learns to reduce errors by predicting the mean depth values of photometric patterns across the dataset. Therefore, averaging single-frame depth predictions at a fixed camera pose will not produce depth estimates that are closer to the truth because of the biased error distribution of depth values. We should not assign high confidence just because the number of frames fused is large. Second, the ability to update the 3D model dynamically as the environment changes is important in endoscopy applications, where the patient anatomy is deformed either by the endoscope itself or by surgical intervention. Even though single-frame depth estimation fits naturally to this scenario, we still need a depth map fusion method that fuses depth information by considering the uncertainty estimates and temporal order to build a dynamic 3D reconstruction. Therefore, the method should update its model rapidly once depth information with high confidence is obtained even if extensive video frames have been fused. The proposed incremental fusion rules are
\begin{equation}
\begin{split}
    &r = \text{max}\left(C_1, \text{min}\left(C_2, \dfrac{\left(S_{i+1}\left(f\left(\mathbf{x}\right)\right)\right)^2}{\left(S_{i+1}\left(f\left(\mathbf{x}\right)\right)\right)^2 + \left(\Sigma_{i}\left(\mathbf{x}\right)\right)^2}\right)\right)
    \\
    &D_{i+1}\left(\mathbf{x}\right) = r D_i\left(\mathbf{x}\right) 
    + \left(1-r\right) d_{i+1}\left(\mathbf{x}\right)
    \\
    &\Sigma_{i+1}\left(\mathbf{x}\right) = r \Sigma_{i}\left(\mathbf{x}\right) + \left(1 - r\right) S_{i+1}\left(f\left(\mathbf{x}\right)\right) \quad \text{, where}
\end{split}
\end{equation}
$\mathbf{x}$ is the spatial location of the volume for depth fusion. $f\left(\mathbf{x}\right)$ is used to find the corresponding location on the image plane of the current fusing frame. $D_i\left(\mathbf{x}\right)$ and $\Sigma_i\left(\mathbf{x}\right)$ are the truncated signed distance and uncertainty estimate at location $\mathbf{x}$ , respectively, obtained by integrating the information from frame $1$ to $i$. $d_{i}\left(\mathbf{x}\right)$ is the truncated signed distance at location $\mathbf{x}$ for frame $i$. $C_1$ and $C_2$ are used to prevent a single fusing frame from having extreme or no impact on the overall reconstruction. The $\delta$ in truncated signed distance function is replaced with the corresponding predicted standard deviation of the fusing frame. Because $\Sigma_i\left(\mathbf{x}\right)$ provides an estimate of isotropic noise models for the fused 3D reconstruction, registration algorithms which assume point-wise noise models~\cite{sinha2018endoscopic} will benefit from that and provide a confidence estimate which is important in clinical applications. The meshes of a 3D reconstruction are generated by applying the Marching Cubes algorithm~\cite{lorensen1987marching} to the implicit surface representation as a signed distance function.\newline
\noindent\textbf{Automatic Failure Detection for Dense 3D Reconstruction:} For clinical applications, reliability and failure-awareness are important. Here, we propose an automatic 3D reconstruction failure detection method. The overall idea is to check if there exists inconsistency among results from SfM, predicted depth maps from the network, and the dense 3D reconstruction. First, consistency should exist between the simulated depth maps from the 3D reconstruction and the predicted depth maps for the same camera poses, if the anatomy does not deform rapidly when these corresponding video frames are captured. Second, pair-wise feature matches and camera poses from SfM should agree with the 3D reconstruction. Third, simulated depth maps of adjacent frames should be consistent and have decent region overlap. These three metrics are calculated using \emph{Dense Simulation Loss}, \emph{Sparse Flow Loss}, and \emph{Depth Consistency Loss}, respectively. In \emph{Sparse Flow Loss} of~\cite{Liu2019Self}, the dense flow map is calculated using simulated depth maps instead of predicted depth maps. In \emph{Depth Consistency Loss}, the predicted mean depth maps are replaced with simulated depth maps. By averaging \emph{Dense Simulation Loss} over all camera poses, \emph{Sparse Flow Loss} and \emph{Depth Consistency Loss} over all pairs of adjacent video frames, we have three metrics. If any of these metrics has an abnormally high value, inconsistency exists and the reconstruction will be treated as failure.\newline
\noindent\textbf{Dense Pose Graph Optimization:} For errors from SfM that cause 3D reconstructions to fail, we need a way to try reducing these errors before rerunning SfM. Even though SfM is robust, there are still extreme cases where there are almost no visual features on tissue surfaces. We use predicted depth maps from \emph{Depth Estimation} and pair-wise feature matches from SfM to optimize the camera pose graph in a differentiable way. Compared to bundle adjustment, this method adds more visual cues in the form of learning-based depth priors apart from the distinguishable textures that SfM solely relies on, which helps to reduce ambiguities when features are sparse. The variables that this method optimizes over are relative coordinate transformations from all the other frames to the first one. All variables are initialized with relative poses from SfM. To exploit pair-wise feature matches from SfM, one component of the objective is \emph{Sparse Flow Loss}. \emph{Depth Consistency Loss}, as another component, is used to exploit geometric constraints among video frames using learning-based depth priors. Because this module only handles cases where camera poses have minor drifting errors, instead of including all pair-wise combinations, we limit the pairing range to neighboring frames for both robustness and computational efficiency. Complete SfM failures are handled by rerunning SfM with different hyper-parameters. Optimized camera poses are used in \emph{Depth Map Fusion} for reconstruction and \emph{Depth Estimation} for network training.
\section{Experiments and Results}
\noindent\textbf{Experimental Design:} The experiments are conducted on a workstation with $4$ NVIDIA Tesla M60 GPUs, each with $8$ GB memory. All modules are implemented using PyTorch. The dataset we train and test on are rectified sinus endoscopic videos ($8$\,min of content in total) acquired from $6$ anonymized and consenting patients with $6$ endoscopes, respectively, under an IRB approved protocol. In a cross-patient study, we conduct $4$ leave-one-out experiments to test on all patients with corresponding CT scans. In each experiment, \emph{Depth Estimation} is trained with video from the remaining $5$ patients, while the resulting 3D reconstructions from the video of the left-out patient are used for evaluation. We did not use a validation set because of the small amount of data. Before 3D reconstructions from \emph{Depth Map Fusion} become available, we train the single-frame depth estimation network with the same architecture, hyper-parameter settings (except for weights of the loss function), and learning scheme as in~\cite{Liu2019Self}. After the pipeline is run once, \emph{Dense Simulation Loss} is incorporated. The weights for \emph{Sparse Depth Loss}, \emph{Depth Consistency Loss}, \emph{Sparse Flow Loss}, and \emph{Dense Simulation Loss} are set to $1.0$, $0.5$, $100.0$, and $0.1$, respectively. $\epsilon$ is $1.0e^{-8}$. We train for $3$ pipeline iterations, each with $20$ epochs. Besides the original data augmentation which includes color shift, blur, and additive noise, we add random flipping and rotation to increase spatial variation. $C_1$ and $C_2$ are $0.1$ and $0.8$. Colors of the reconstructions are fused using the same exponential averaging method as the signed distance. The thresholds in \emph{Failure Detection} are $2.0$, $0.1$, and $2.0$ for \emph{Dense Simulation Loss}, \emph{Sparse Flow Loss}, and \emph{Depth Consistency Loss}, respectively. The frame interval for the second and third metric calculation is $5$. The weights in \emph{Pose Graph Optimization} are $1.0$ for \emph{Depth Consistency Loss} and $100.0$ for \emph{Sparse Flow Loss}. The frame intervals are set to $5$,$6$,$7$, and $8$. For evaluation, only 3D reconstructions that are determined to be successful are used. 3D reconstructions from \emph{Depth Map Fusion} are converted from meshes to point clouds and uniformly downsampled to around $40k$ points before evaluation. We rigidly register the point clouds to the corresponding patient CT mesh models and use the residual errors from the registration algorithm~\cite{billings2015generalized} as the accuracy of our dense reconstructions.

\begin{figure}[t]
	\centering
	\includegraphics[width=120mm]{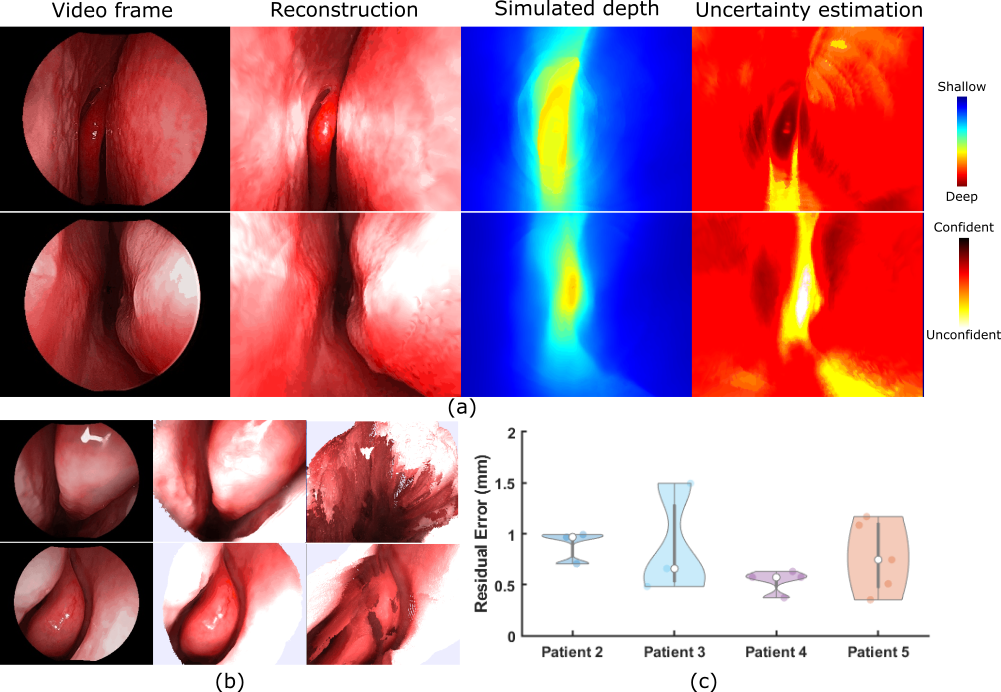}
	\caption{Qualitative and quantitative results. Readers are recommended to watch the supplementary video for better visualization. (a) From left to right: original video frame; color rendering, simulated depth map, and uncertainty estimate of the 3D reconstruction from the same pose. (b) From left to right: original video frame, 3D reconstruction after and before pose optimization. Failures are detected using \emph{Failure Detection}. (c) Residual errors of reconstructions from testing videos. All errors are below $1.5$\,mm.} 
	\label{fig:results}
\end{figure}

\noindent\textbf{Results:} Qualitative and quantitative results are shown in Fig.~\ref{fig:results}. Most of the reconstructions are generated using video sequences longer than $20$ seconds. Compared to single-frame methods, the residual errors of our reconstructions better represent the true accuracy because of the larger coverage of anatomy in each reconstruction.

\section{Discussion and Conclusion}
In this work, we present a self-supervised approach for dense 3D reconstruction from full-length monocular endoscopic videos with uncertainty estimation, failure detection, and error recovery. Our method, in principle, is able to generate and update the 3D reconstructions on the fly in the application phase if SfM is replaced with a SLAM-based method. We do not train a network to predict relative camera poses because the inherent difficulty of global scale estimation from monocular endoscopic images makes the task infeasible without additional prior input. The dense 3D reconstruction provided by our method is accurate up to a global scale and we rely on other information or devices, such as CT scans or inertial measurement units, to recover global scale. In future work, we will investigate self-supervised dynamic 3D reconstruction on the fly.

\bibliography{main}

\begin{thebibliography}{10}

\bibitem{sinha2018endoscopic}
Sinha, A., Liu, X., Reiter, A., Ishii, M., Hager, G.D., Taylor, R.H.:
\newblock Endoscopic navigation in the absence of ct imaging.
\newblock In: MICCAI 2018, Cham, Springer International Publishing (2018)
  64--71

\bibitem{leonard2018evaluation}
{Leonard}, S., {Sinha}, A., {Reiter}, A., {Ishii}, M., {Gallia}, G.L.,
  {Taylor}, R.H., {Hager}, G.D.:
\newblock Evaluation and stability analysis of video-based navigation system
  for functional endoscopic sinus surgery on in vivo clinical data.
\newblock IEEE T MED IMAGING \textbf{37}(10) (Oct 2018)  2185--2195

\bibitem{grasa2014visual}
Grasa, O.G., Bernal, E., Casado, S., Gil, I., Montiel, J.:
\newblock Visual slam for handheld monocular endoscope.
\newblock IEEE T MED IMAGING \textbf{33}(1) (2014)  135--146

\bibitem{mahmoud2017slam}
Mahmoud, N., Hostettler, A., Collins, T., Soler, L., Doignon, C., Montiel, J.:
\newblock Slam based quasi dense reconstruction for minimally invasive surgery
  scenes.
\newblock arXiv preprint arXiv:1705.09107 (2017)

\bibitem{goncalves2015perspective}
Goncalves, N., Roxo, D., Barreto, J., Rodrigues, P.:
\newblock Perspective shape from shading for wide-fov near-lighting endoscopes.
\newblock Neurocomputing \textbf{150} (2015)  136--146

\bibitem{yang2016compact}
Yang, S.P., Kim, J.J., Jang, K.W., Song, W.K., Jeong, K.H.:
\newblock Compact stereo endoscopic camera using microprism arrays.
\newblock Opt. Lett. \textbf{41}(6) (2016)  1285--1288

\bibitem{mahmood2018deep}
Mahmood, F., Durr, N.J.:
\newblock Deep learning and conditional random fields-based depth estimation
  and topographical reconstruction from conventional endoscopy.
\newblock MED IMAGE ANAL (2018)

\bibitem{Liu2019Self}
{Liu}, X., {Sinha}, A., {Ishii}, M., {Hager}, G.D., {Reiter}, A., {Taylor},
  R.H., {Unberath}, M.:
\newblock {Self-supervised Learning for Dense Depth Estimation in Monocular
  Endoscopy}.
\newblock arXiv e-prints (Feb 2019)  arXiv:1902.07766

\bibitem{kendall2018multi}
Kendall, A., Gal, Y., Cipolla, R.:
\newblock Multi-task learning using uncertainty to weigh losses for scene
  geometry and semantics.
\newblock In: CVPR. (2018)  7482--7491

\bibitem{curless1996volumetric}
Curless, B., Levoy, M.:
\newblock A volumetric method for building complex models from range images.
\newblock (1996)

\bibitem{zach2007globally}
Zach, C., Pock, T., Bischof, H.:
\newblock A globally optimal algorithm for robust tv-l 1 range image
  integration.
\newblock In: ICCV, IEEE (2007)  1--8

\bibitem{lorensen1987marching}
Lorensen, W.E., Cline, H.E.:
\newblock Marching cubes: A high resolution 3d surface construction algorithm.
\newblock In: ACM siggraph computer graphics. Volume~21., ACM (1987)  163--169

\bibitem{billings2015generalized}
Billings, S., Taylor, R.:
\newblock Generalized iterative most likely oriented-point (g-imlop)
  registration.
\newblock IJCARS \textbf{10}(8) (2015)  1213--1226

\end{thebibliography}

\end{document}